\pdfoutput=1
\documentclass[10pt,letterpaper]{article}

\usepackage{cogsci}

\cogscifinalcopy 

\usepackage{graphicx}
\usepackage{courier}
\usepackage{amsmath}
\usepackage{amsfonts}

\usepackage{pslatex}
\usepackage{apacite}
\usepackage{float} 

\title{Measuring and Modeling Physical Intrinsic Motivation}
 
\author{{\large \bf Julio Martinez\textnormal{\textsuperscript{1,$^\ast$}}, Felix Binder\textnormal{\textsuperscript{2}}, Haoliang Wang\textnormal{\textsuperscript{3}}, Nick Haber\textnormal{\textsuperscript{4,5}}, Judith Fan\textnormal{\textsuperscript{1,3}}, Daniel L. K. Yamins\textnormal{\textsuperscript{1,5}}} \\
  \textsuperscript{1}Department of Psychology, Stanford University\\
  \textsuperscript{2}Department of Cognitive Science, University of California San Diego\\
  \textsuperscript{3}Department of Psychology, University of California San Diego\\
  \textsuperscript{4}Graduate School of Education, Stanford University \\
  \textsuperscript{5}Department of Computer Science, Stanford University\\
  \textsuperscript{$^\ast$}juliomz@stanford.edu}

\begin{document}

\maketitle

\begin{abstract}
Humans are interactive agents driven to seek out situations with interesting physical dynamics. Here we formalize the functional form of physical intrinsic motivation. We first collect ratings of how interesting humans find a variety of physics scenarios. We then model human interestingness responses by implementing various hypotheses of intrinsic motivation including models that rely on simple scene features to models that depend on forward physics prediction. We find that the single best predictor of human responses is adversarial reward, a model derived from physical prediction loss. We also find that simple scene feature models do not generalize their prediction of human responses across all scenarios. Finally, linearly combining the adversarial model with the number of collisions in a scene leads to the greatest improvement in predictivity of human responses, suggesting humans are driven towards scenarios that result in high information gain and physical activity.

\end{abstract}

\section{Introduction} From infancy, humans exhibit strong intrinsic motivations to curiously explore their physical environments and play with the objects they encounter. Their exploration patterns are not random but follow an underlying systematic approach. More generally, curiosity has been described as an intrinsic motivation that aids in closing gaps in knowledge \cite{loewenstein1994psychology, oudeyer2009intrinsic, kidd2015psychology}. Many works have contributed to the understanding of human preferences during free-play and self-directed learning. Previous work shows that children prefer stimuli of intermediate predictivity \cite{cubit2021visual} and that the inability to predict an outcome of an action is a powerful driver that steers information seeking behavior \cite{markant2014preference}. Self-directed learning allows humans to focus their efforts on useful information they do not yet posses resulting in highly selective but efficient information sampling strategies \cite{kidd2012goldilocks}.

A core component of physical intrinsic motivation is how it influences the learning of a physical world model (WM). In learning physical WMs infants are faced with the need to observe a wide range of complex physical dynamics \cite{kim2020active}. Thus it is a natural hypothesis that in order to gather data to learn good WMs there must be a powerful exploration strategy. 

In this work we seek to formalize physical intrinsic motivation -- the problem of determining the \textit{intrinsic reward function} (IRF) that directs exploration actions in humans during free-play in physics environments. To do this we directly probe ``interestingness" as a proxy for intrinsic reward (IR) and evaluate various IRFs as conceptually distinct hypotheses. Our contributions are as follows: 1. We collect human interestingness ratings over a diverse range of 3D simulated videos. 2. We formulate the formal theory of intrinsically motivated agents with different IRFs. 3. We compare different IRFs and composites thereof in terms of their predictivity of human ratings and their generalization across scenario types.

Our core conclusions are that WM-based IRFs generalize better than simple scene features in predicting human responses across scenarios suggesting a promising direction to improve the current set of WM-based IRFs. Second, that humans are characterized in part by both information seeking and high activity seeking motivations -- humans find as interesting the stimuli whose outcomes are hard to predict or fun to watch because of the many collisions observed. 

\section{Theoretical Framework} 
\begin{figure*}[ht]
    \centering
        \includegraphics[width=\textwidth]{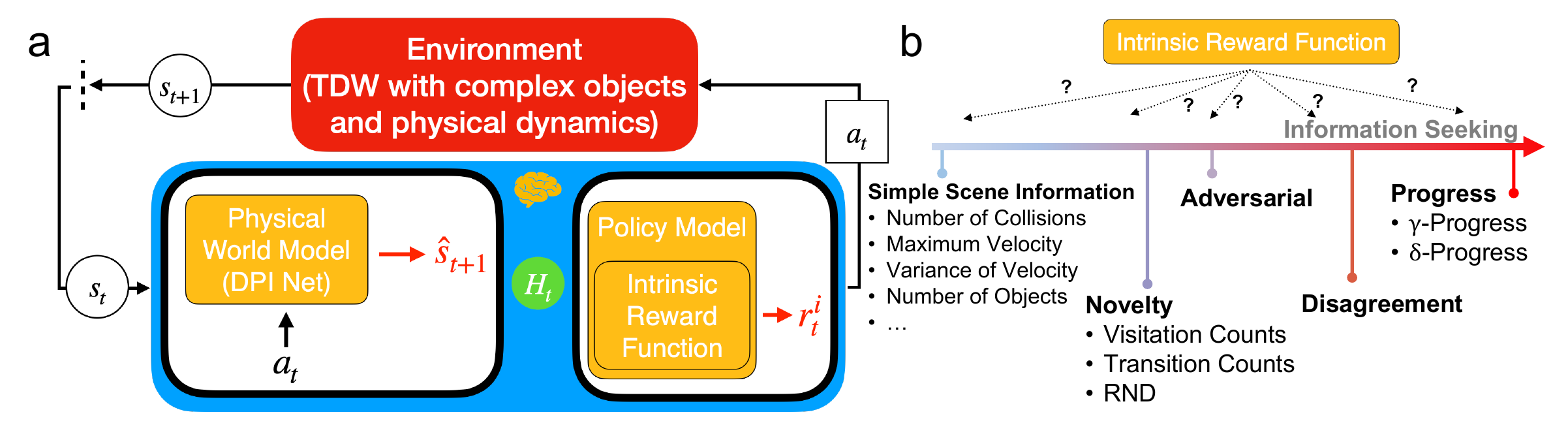}
    \caption{\textbf{Theoretical framework for an IR agent.} \textbf{(a)} Intrinsic reward RL agent acting in a simulated environment (TDW) that uses a WM for forward physics prediction, and a PM that optimizes an IR generated by an IRF. \textbf{(b)} A graphic describing the hypothesis space for IRFs, in the form of a continuum that progressively becomes more explicitly information seeking towards the right of the continuum. } 
    \label{fig:figure1}
\end{figure*}

\subsubsection{Intrinsically Motivated Agents}
We start by using a Reinforcement Learning (RL) framework as in \cite{haber2018learning, kim2020active} to define an intrinsically motivated agent who generates its own IR (Fig.~\ref{fig:figure1}a). At time $t$ an RL agent makes an observation from state $s_t \in S$ and takes an action $a_t \in A$ which results in a state transition $s_{t+1} \in S$ (i.e. the physical dynamics). We say the agent is intrinsically motivated when it's IR $r_t$ is based on its own observations of the environment or its predictions of state transitions.  

The Policy Model (PM) is a function $PM: S \rightarrow A \label{eq:policymodel}$ that maps the state to an action. The PM is built around a subcomponent called the IRF that generates the IRs. Given a history $H_t$ of states, actions, and IRs, the agent is able to learn a policy that optimizes for total IR. The physical WM, parameterized by $\theta_{WM}$ (i.e. the agent's knowledge of the world), allows the agent to simulate forward in time the physical dynamics of its environment. Its functional form takes as input the state of the environment and an action, and predicts the next state $WM: S \times A \rightarrow S$. The IRF is a function that generates the total reward from a sequence of state action pairs and can potentially depend on the state of the environment, the action being taken, and the state of the agent's knowledge of how the world evolves over time. 
\begin{align*}
    &IRF: S \times A \times \theta_{WM} \rightarrow \mathbb{R} 
\end{align*}

\subsubsection{Intrinsic Reward Functions}
We conceptualize the hypothesis space for IRFs by the degree to which they explicitly improve the WM i.e. by an information seeking continuum (Fig. \ref{fig:figure1}b). On the left extreme of this continuum we have simple scene features, such as the number of object collisions, a data gathering signal which may or may not result in improving the WM. Progressing towards the right we have IRF classes that are increasingly designed to explicitly improve the WM. The \textit{novelty} of the state \cite{tang2017study} generates more reward for unfamiliar states effectively improving the WM by providing diverse training stimuli. \textit{Adversarial} generates rewards proportional to the WM loss \cite{stadie2015incentivizing, pathak2017curiosity}. \textit{Disagreement} generates rewards based on the variance in predictions from multiple WMs \cite{pathak2019self} to minimize uncertainty across WM predictions. At the far right we have \textit{Progress} which assigns higher reward to observations that reduce the WM loss \cite{ten2021humans, schmidhuber2010formal, graves2017automated, achiam2017surprise}. It is important to note that specific implementations of each IRF vary and may not preserve the order in which their conceptual counterpart appears on the continuum. To be explicit, we give details of the specific IRF implementations used for this work below. 


 \textit{Simple Scene Features} based on the information readily available in the scene of each stimulus, as in \cite{curiodrop, curiotower}, were used as candidate IRFs. These IRFs do not depend on a physical WM. They are defined as follows: \textit{Position}: the x, y and z coordinates of all positions for all objects in a scene from which we compute the mean, variance, min, and max. \textit{Velocity}: the x, y, and z coordinates of the velocities of all objects in a scene from which we compute the mean, variance, min, and max. \textit{Variance of position}: the trace of the covariance of all object’s positions from which we compute the initial value, mean, min, and max over several stimulus frames. \textit{Variance of velocity}: the trace of the covariance of all object's velocities from which we compute the mean, min, max and initial value. \textit{Number of collisions}: the number of collisions occurring over an entire stimulus. We also compute the initial number of collisions (how many objects are in contact that separate initially), mean, min, and max number of collisions over temporal intervals throughout the stimulus. \textit{Number of objects}: the number of objects in a stimulus. \textit{Number of object categories}: the total number of different types of objects in a stimulus. \textit{Number of distractors}: total number of objects that are fixed and that do not interact with other objects.

 \begin{figure*}[ht]
    \centering
        \includegraphics[width=\textwidth]{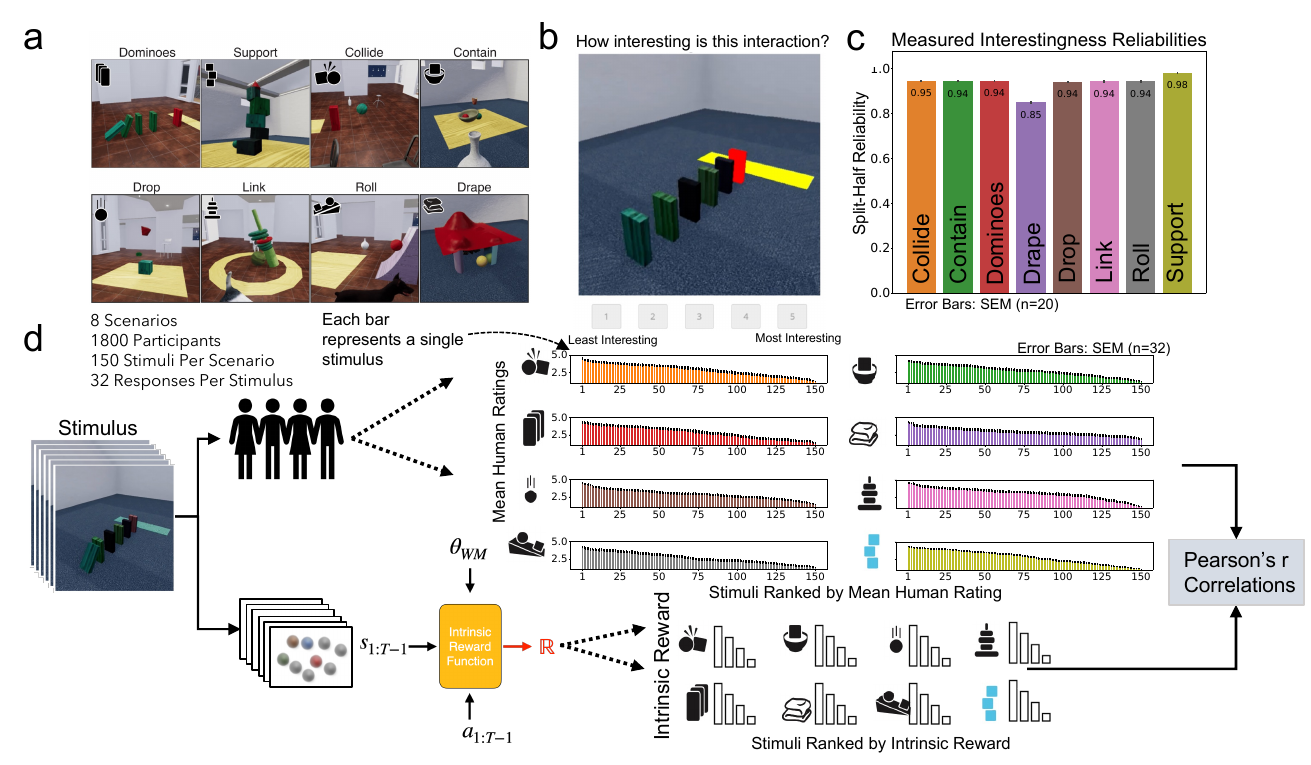}
    \caption{ \textbf{Stimuli, Task Procedure and Modeling Pipeline.} \textbf{(a)} Stimulus examples of the 8 scenario categories from the Physion Benchmark dataset. Stimuli contain a diverse set of rigid and non-ridge body physical dynamics useful for probing and modeling physical intrinsic motivation. \textbf{(b)} In the experimental task, participants are shown the stimulus videos. After watching participants are asked to rate how interesting they find the video on a Likert scale from 1 to 5 (5 being most interesting). \textbf{(c)} The scenario mean split-half reliabilities for human ratings (n=20 split-halves for each stimulus and averaged across stimuli in each scenario). \textbf{(d)} The experimental and modeling pipeline that feeds stimulus to both humans and various IRFs. Mean human responses for each stimulus are correlated to IRF values to compare their predictivity on human interestingness.
    } 
    \label{fig:figure2}
\end{figure*}

\textit{Random Network Distillation} (RND) is a novelty based IRF \cite{burda2018exploration}. In RND a randomly initialized target network, $\mu_{\theta_{target}}$ takes as input the state at time $t$, $s_{t} \in S$, and embeds it into a d-dimensional representation. A predictor network, $\mu_{\theta_{predictor}}$, randomly initialized using a different seed, is trained to minimize the RMSE between its prediction of the target network's embeddings. IR for a single state observation is equal to the RMSE between the predicted and target embeddings. RND does not rely on WM prediction, but does rely on the predictor network learning of the target network embeddings. 
\begin{align*}
    r_{t, rnd}^i &= RMSE(\mu_{\theta_{target}}(s_t), \mu_{\theta_{predictor}}(s_t))
\end{align*}

\textit{Adversarial} reward, an approximation to surprisal \cite{achiam2017surprise} sets the IR for a single step equal to the WM loss for a $k$-step rollout. The $k$-step rollout is an autoregressive rollout where the WM first takes as input the current state $s_t$ and predicts $\hat{s}_{t+1}$ (one step ahead). For a 2-step rollout, the WM feeds its 1-step prediction, $\hat{s}_{t+1}$ as input to make a 2-step rollout prediction $\hat{s}_{t+2}$, and so forth for a $k$-step rollout, $\hat{s}_{t+k}$. The MSE between the ground truth state at the $k$th step,$s_{t+k}$ and the predicted $\hat{s}_{t+k}$ is the IR value. 
\begin{align*}
    r_{t, adversarial}^i &= \mathcal{L}_{ \theta_{WM}}(s_{t+k}, \hat{s}_{t+k}) = MSE(s_{t+k}, \hat{s}_{t+k})
\end{align*}

\begin{figure*}[ht]
    \centering
        \includegraphics[width=\textwidth]{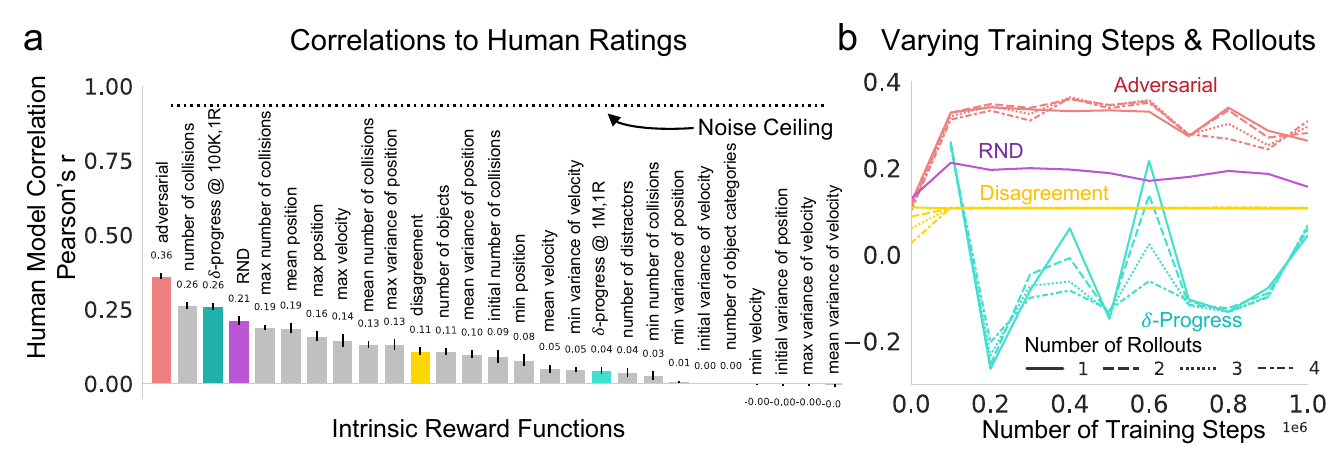}
    \caption{ \textbf{Model-Human Correlation Results for Individual IRF Candidates.}  \textbf{(a)} The correlations of IRFs to mean human ratings. Adversarial reward has the highest correlation to humans. $\delta$-progress is shown twice, first at $0.1 \times 10^6$ training steps of DPINet and again at $1 \times 10^6$ training steps for DPINet. \textbf{(b)} We vary the amount of pretraining of our WM by varying the number of training steps for DPINet that adversarial, disagreement, and $\delta$-progress depend on for WM prediction. We also vary the number of rollout steps DPINet makes for its predictions in each pretrained DPINet. Like DPINet we also vary the number of training steps for the RND predictor network. RND however does not depend on rollouts. As we vary the training, adversarial, disagreement, and RND are relatively stable after $0.1 \times 10^6$ time steps. $\delta$-progress on the other hand is relatively unstable often changing from positive to negative correlation over many time steps. Adversarial, disagreement, and $\delta$-progress do not change much as a function of the number of rollouts. } 
    \label{fig:figure3}
\end{figure*}

\textit{Disagreement} \cite{pathak2019self} depends on multiple WMs each of which generates $k$-step rollouts resulting in predictions $\hat{s}_{t+k}^{\theta_{WM_1}}$, $\hat{s}_{t+k}^{\theta_{WM_2}}$, and $\hat{s}_{t+k}^{\theta_{WM_2}}$. The IR at time $t$ is assigned the mean variance across each WM's $k$-step predictions  made from time $t$.
\begin{align*}
    r_{t, disagreement}^i &= Mean(Var(\hat{s}_{t+k}^{\theta_{WM_1}},\hat{s}_{t+k}^{\theta_{WM_2}}, \hat{s}_{t+k}^{\theta_{WM_3}}))
\end{align*}

\textit{$\delta$-Progress} computes the difference between an old and new WM's loss of a $k$-step rollout prediction as shown below \cite{graves2017automated, achiam2017surprise}. The new WM is the current WM, or more generally, the WM trained after $n$ number of training iterations parameterized by $\theta_{WM_{n}}$. The old WM is $\delta$ training steps back, parameterized by $\theta_{WM_{n-\delta}}$.
\begin{align*}\label{eq:progress}
    &r_{t, progress}^i = \mathcal{L}_{\theta_{WM_{n-\delta}}}(s_{t+k}, \hat{s}_{t+k}) - \mathcal{L}_{\theta_{WM_{n}}}(s_{t+k}, \hat{s}_{t+k}) 
\end{align*}

Each IRF intermediately outputs an IR $r_t^i$ for every transition in the environment. The IRF sums over the IRs of an entire stimulus of length $T$ to compute the total IR, $\sum_{t=0}^{T-1}r_t^i \in \mathbb{R}$.

\subsubsection{Choice of World Model} In real life humans do not have access to the exact state of the world and instead infer it from perception through partial observation. Here we bypass perception, as it is out of scope for the problem under investigation, and directly feed our WMs an explicit state description as input. The state description is generated by converting the 3D state of the virtual environment to a 3D mesh from which object centric coordinates or particles are used to describe each object's position and velocity in the scene. 

For physics prediction, we chose a model whose performance best represents human prediction accuracy, DPINet \cite{dpinet}. DPINet is part of a class of neural network scene graph physics models that learn physical prediction via local and hierarchical back-propagation \cite{dpinet, Mrowca2018}. DPINet showed the most human like accuracy and consistency on Physion when compared to other earlier versions of scene graph based models and several image based models, see Fig. 5 in \cite{bear2021physion}. DPINet was pretrained on the Physion training set to reproduce the performance results previously stated \cite{bear2021physion}. DPINet receives as input the state particle description, the positions and velocities of all the object particles, and makes a forward prediction of the positions and velocities of all object particles for the next time step. 

\begin{figure*}[ht]
\centering
\includegraphics[width=\textwidth]{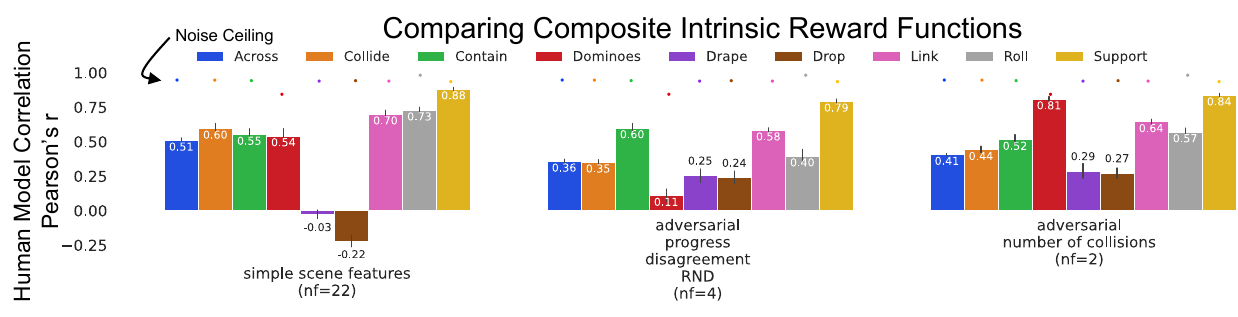}
\caption{\textbf{Analysis of IRF Composites.} 
Three groups of bar plots correspond to three composite functions of IR. The first group on the left is a composite model using all simple scene features as predictors in linear regression. While correlation performance is 0.51 across scenarios, as indicated by the blue bar, Drop and Drape scenarios have small and negative correlation indicating an inability for linear combinations to generalize across scenarios well. The center group shows the results if a composite model of all WM-based IRFs and RND. Generalization is much better across scenarios although still below noise ceiling. The third group on the right shows a composite model of the adversarial IRF with number of collisions, the feature that when linearly combined with adversarial improves accuracy across scenarios the most.} 
\label{fig:figure4}
\end{figure*}

\section{Experiment}
For our experiment we gathered human interestingness ratings on a wide range of physics scenarios to serve as a target in modeling IR. 

\subsubsection{Participants} We recruited 1,800 participants from Prolific to complete the task. Participants provided informed consent and were paid approximately \$14 per hour. 

\subsubsection{Stimuli} We use the Physion Benchmark dataset \cite{bear2021physion} generated from 3D simulated videos in ThreeDWorld \cite{gan2020threedworld}. We repurpose Physion by changing the design variable to ``interestingness"  (instead of prediction accuracy) as a proxy for total IR by asking adult humans how interesting they find the stimulus. We chose these stimuli because they provide a diverse set of rigid and non-rigid body dynamics wherein to study physical intrinsic motivation. Physion is split into a training set, used to train WMs and IRFs, and a test for the experiment and for evaluating IRFs. Each set contains 8 scenarios: Collide, Contain, Dominoes, Drape, Drop, Link, Roll, and Support shown in Fig.~\ref{fig:figure2}a. There are 2000 and 150 videos for each scenario in the train and test sets respectively. 

\subsubsection{Task Procedure} Participants were asked for interestingness ratings (Fig~\ref{fig:figure2}b) after observing the outcomes in each stimulus video in the Physion test set. Each of 150 trials began with a fixation cross, shown for a randomly sampled time between 500ms and 1500ms. Participants were then shown the first frame of the video for 2000ms after which the entire video was played. Once the video stopped playing, it was removed and the response buttons were enabled. The experiment moved to the next phase after participants selected an interestingness rating from 1 to 5. Participants were presented with the stimuli in a randomized sequence and were only allowed to take the task once. Data was excluded from participants that did not complete trials, whose exit survey indicated they did not understand the study, or failed to include at least one response for each endpoint of the scale.

\subsubsection{Validation and Reliability} We report the reliability between participants (mean of n=32 responses per stimulus) as the split-half reliability using Spearman Brown correction. We compute the average split-half reliability for a given stimulus using 20 split halves and average across stimulus within each scenario category. The cross scenario category mean split-half reliability was 0.935 with 0.002 SEM. All scenario category reliabilities are shown in shown in Fig.~\ref{fig:figure2}c. 

\subsubsection{Data Analysis Pipeline} Comparing human action choices during free-play to the action choices of an RL agent is a natural modeling choice. However, doing so requires a PM to plan the optimal sequence of actions. Incorrectly choosing a suboptimal PM introduces an additional source of error in planning. We avoid this confound by using IRFs to instead directly generate total IR for our predefined stimulus and forgo using a PM for interactive RL. We compute Pearson r correlations to evaluate how predictive each IRF is to mean human ratings for each of the 150 rated stimuli in all 8 scenarios depicted in the upper half of Fig.~\ref{fig:figure2}d. \\
\indent To compute IR from WM-based IRFs we pretrain DPINet for 1 million time steps on the training stimulus set from Physion, resulting in similar loss from \cite{bear2021physion}. Before training we extracted x, y, z particles representations for object positions and velocities at each frame of the stimuli. These object particle representations are fed as input into DPINet and the IRFs. We compute $\delta$-progress, disagreement, and adversarial total reward at several training steps of DPINet $(0, 100\times10^3, 200\times10^3, ..., 1\times10^6)$. For each training step of DPINet, the WM-based IRFs were computed based on 1, 2, 3, or 4-step rollouts to test whether simulating further forward in time resulted in significant differences in IR prediction. For disagreement we pretrained DPINet models with 3 initialization seeds. For $\delta$-progress $\delta=$50,000 gave best results over a grid search on $\delta$. For RND, we modified only the final layer of DPINet to output a 200 dimensional state embedding. This modified DPINet served as the architecture for our RND predictor and target networks (both architectures were identical with different initialization seeds). The predictor network was pretrained for the same number of time steps as DPINet. \\
\indent In addition to evaluating individual IRFs, we trained linear regression models to predict human ratings using L1 regularization and leave-one-out cross validation to find the best fits for composite IRFs. We evaluated all model predictions by correlating them to mean human ratings. 10 random splits from the stimulus set were drawn and for each split 80\% was used as regression model training data. The remainder 20\% stimuli was used for evaluating predictions. The same test splits were used to evaluate the individual IRFs. 

\section{Results}
\begin{figure*}[ht]
\centering
\includegraphics[width=0.95\textwidth]{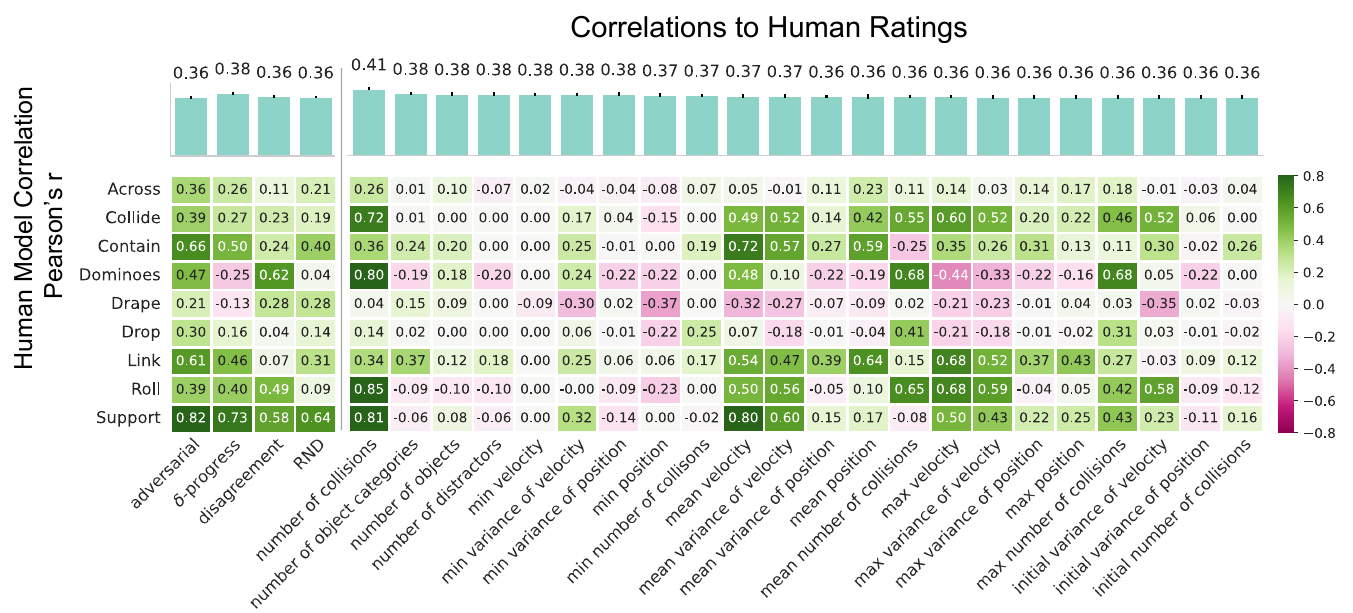}
\caption{\textbf{Bottom Panel: Investigating per scenario results for each candidate IRF.} Each box represents the correlation of the indicated IRF (columns) for the indicated scenario (rows). All green within a column indicates the given IRF has the same sign across all scenarios; Columns mixed with green and pink indicate inconsistent correlation signs across scenarios and thus lack of generalization.\textbf{Top Panel: Combining adversarial IRF with other IRFs.} Each blue bar represents the predictive accuracy of the two-feature model linearly combining the best single-feature IRF (adversarial) with the indicated feature. Bars are ordered by complementarity with adversarial.} 
\label{fig:figure5}
\end{figure*}

\textbf{Explicitly world-model-based IRFs tend to better explain human interestingness ratings than simple scene features, with adversarial loss achieving the best overall match.} Evaluating the predictivity of single IRFs across all scenario types (Fig. \ref{fig:figure3}a), we found a range of predictivity levels, with several simple scene features indistinguishable from 0 in their predictivity. In contrast, explicitly WM-based IRFs included 3 of the top 4 best predictors. Adversarial loss achieved the best predictivity, with an average cross-scenario correlation of 0.360 (~0.39\% of noise ceiling of 0.935). 

\textbf{World-model-based intrinsic reward functions are stable across rollouts and, mostly, across training steps.} As shown in Fig.~\ref{fig:figure3}b, correlations of all IRF models were stable with respect to rollout length. Adversarial, progress, and RND provide stable correlations to human ratings after $100 \times 10^3$ training steps, with relative rankings maintained throughout. $\delta$-Progress IRF was an exception, changing significantly from positive to negative correlation across training steps. (For this reason, Fig.~\ref{fig:figure3}a shows $\delta$-progress results at both the maximally-correlated and final timesteps.) All IRF models included here are based on DPINet (or modified DPINet for RND). In future work we will measure the effects of using other WM architectures including pixel-based forward predictors such as TECO \cite{yan2022temporally} and MCVD \cite{voleti2022masked}. 

\textbf{Simple scene features do not generalize across scenarios but world model based intrinsic reward functions do.} Linearly combining groups of simple scene features as predictors improves overall predictivity of human ratings, but reduces predictivity in specific scenarios, particularly Drop and Drape (Fig.~\ref{fig:figure4}, left). Interestingly, this outcome reproduces previous work from \cite{curiotower, curiodrop} where the Support scenario was predicted well using scene features, but the Drop scenario was not. Looking at the relationships of individual simple scene features to each scenario, we find that the cause of this outcome is the change of sign in relationship across scenarios (columns of Fig.~\ref{fig:figure5}). This indicates that, to the extent that simple scene features ``explain'' interestingness, they do so in a highly scenario-type specific manner, and do not generalize.
On the other hand, inspecting per-scenario results from the explicitly WM-based IRFs shows that the explanatory direction is consistent across all scenarios (Fig.~\ref{fig:figure4}, center),
indicating improved generalization. 
Correlation signs are consistently positive across scenarios for all world-model-based IRFs (Fig.~\ref{fig:figure5}).\\
\indent \textbf{All intrinsic reward function correlations to human ratings are far below noise ceiling.} Despite some explanatory power in several IRF candidate models, all are far below noise ceiling both across scenarios (dashed line in Fig.~\ref{fig:figure3}a) and for most individual scenarios (dots over each column in Fig.~\ref{fig:figure4}). \\
\indent \textbf{Adversarial reward and number of collisions are the most complimentary intrinsic reward components.} To determine complementarity among our IRF candidates, we built linear combinations of pairs of IRFs. Starting with the best matching adversarial IRF, we found that the most complementary feature was a WM independent IRF, the number of collisions (Fig.~\ref{fig:figure5}, blue bars).
\section{Discussion} 
In this work, we measured human interestingness judgements and assessed several IRFs on their predictivity of human ratings. We observed that IRFs explicitly involving assessment of WM knowledge tended to be better predictors of human data than simple scene features, with adversarial loss having the best overall correlation. This suggests humans are at least partly motivated by ``information seeking'' goals in our experimental setting. Explicit WM-based IRFs also generalized across different types of physical scenarios, while simple scene features did not. Perhaps most saliently, all IRFs tested in this work are well below the human experimental noise ceiling. This encourages us to make further improvements for WM-based IRFs that can account for intrinsic motivation over a wide range of physical scenarios. 

While WM-based IRFs add complexity in modeling forward dynamics, simple scene features require pre-specifying only the relevant features for any new scenario a human might encounter. However, we also observed that the most complementary predictor to adversarial reward was the raw number of collisions in a given scenario, particularly in scenarios like Dominoes where object collisions are especially salient. This suggests participants sometimes valued scenarios as interesting based on the amount of activity rather than explicitly for information gain.  We plan to explore simple scene features that more directly relate to physical activity and that may potentially yield further performance improvements when integrated into a WM-based IRF.  

Finally, we also seek to further investigate if this bifurcation in explanatory modes (i.e. the level of physical activity vs information gain) is an artifact of our experimental design (e.g. because it measures interestingness in non-interactive displays rather than real interactive settings), or a core feature of intrinsic motivation that a better IRF model will need to explain.

\newpage

\section{Acknowledgments}
This work was supported by the following grants: Simons Foundation grant 543061 (D.L.K.Y),
National Science Foundation CAREER grant 1844724 (D.L.K.Y), Office of Naval Research grant S5122 (D.L.K.Y.), Stanford University Human-Centered Artificial Intelligence Inaugural Fellowship.

\bibliographystyle{apacite}
\setlength{\bibleftmargin}{.125in}
\setlength{\bibindent}{-\bibleftmargin}

\bibliography{cogsci_full_paper_template}

\end{document}